# TreeSegNet: Adaptive Tree CNNs for Subdecimeter Aerial Image Segmentation

Kai Yue, Lei Yang, Ruirui Li*, Wei Hu, Fan Zhang, Wei Li

Beijing University of Chemical Technology, Beijing, China

College of Information Science & Technology

North Third Ring Road 15, Chaoyang District, Beijing100029, China

## Abstract

For the task of subdecimeter aerial imagery segmentation, fine-grained semantic segmentation results are usually difficult to obtain because of complex remote sensing content and optical conditions. Recently, convolutional neural networks (CNNs) have shown outstanding performance on this task. Although many deep neural network structures and techniques have been applied to improve the accuracy, few have paid attention to better differentiating the easily confused classes. In this paper, we propose TreeSegNet which adopts an adaptive network to increase the classification rate at the pixelwise level. Specifically, based on the infrastructure of DeepUNet, a Tree-CNN block in which each node represents a ResNeXt unit is constructed adaptively according to the confusion matrix and the proposed TreeCutting algorithm. By transporting feature maps through concatenating connections, the Tree-CNN block fuses multiscale features and learns best weights for the model. In experiments on the ISPRS 2D semantic labeling Potsdam dataset, the results obtained by TreeSegNet are better than those of other published state-of-the-art methods. Detailed comparison and analysis show that the improvement brought by the adaptive Tree-CNN block is significant.

---

* Corresponding author. Tel.: +86 18911938368. E-mail address: ilydouble@gmail.com

*Keywords*: aerial imagery; semantic segmentation; tree structures; adaptive network; ISPRS; CNN

# 1. Introduction

Semantic segmentation is the task of assigning a semantic label (land-cover or land-use class) to every pixel of an image. Recently, highly developed remote sensing techniques have been able to provide very-high-resolution (VHR) aerial images with a ground sampling distance (GSD) of 5-10cm in the spatial or spectral domain. As a result, small objects such as cars and buildings are distinguishable and can be segmented. When processing ultra-high resolution data, most of the previous methods have relied on supervised classifiers that were trained on hand-crafted feature sets, which describe locally the image content. The extracted high-dimensional representation is assumed to contain sufficient information to classify a pixel. In fact, these features depend on a specific feature extraction method whose parameters and performance on the specific data were previously unknown.

Deep convolutional neural networks (DCNNs) have become extremely successful in many high-level computer vision tasks, which range from image classification (Krizhevsky et al., 2012) to object detection, visual recognition, and semantic segmentation. DCNNs address the trainable tasks in an end-to-end fashion, which usually means learning jointly serious feature extraction from raw input data to a final, task-specific output. DCNNs have also been applied to the remote sensing field. The fully convolutional network(FCN) (Long et al., 2015) is a classic DCNN that is designed for semantic segmentation. Sherrah et al. (Sherrah, 2016) applied FCN to remote sensing imagery and published their results on the ISPRS Potsdam benchmark. Extensions of FCNs, including DeconvNet (Noh et al., 2015a), SegNet (Badrinarayanan et al., 2017) and U-Net (Ronneberger et al., 2015) are then used on aerial imagery labeling (Doxani et al., 2012; Liu et al., 2017; Volpi and Tuia, 2017) to obtain higher accuracy.

As often observed, aerial imagery is characterized by complex data properties in the form of heterogeneity and easily mixed classes. Even for the existing well-performed DCNN network structures, semantic labeling on aerial imagery is still difficult and asks for better comprehension of the image context. This paper proposes the TreeSegNet architecture which contains an automatically constructed Tree-CNN block. The Tree-CNN block combined with new short connections are designed for multiclass labeling task of easily confused categories. The TreeSegNet extends the DeepUNet (Li et al., 2018) and achieves an impressive promotion of the accuracy. In summary, we made the following contributions:

1) An end-to-end deep neural network architecture that connects DeepUNet and the tree-CNN block is proposed, which produces better performance on the ISPRS Potsdam dataset than other published methods.

2) This paper introduces an automatic method for constructing the adaptive tree-CNN block. The learned tree-CNN network structure can help to differentiate easily confused classes. As far as we know, it is the first CNN network for semantic labeling tasks that has been adapted to the remote sensing field.

3) A novel modified DeepUNet is introduced for semantic labeling of the VHR remote sensing images, which adopts the ResNeXt concept.

4) A group of experiments are performed on the ISPRS Potsdam dataset, making comparisons among U-Net, DeepUNet, and TreeSegNet. The results are submitted to the ISPRS community.

The remainder of this paper is organized as follows. In Section 2, we will introduce those research topics that are related to the subjects of remote sensing semantic labeling, deep learning, and adaptive networks. The method proposed in this paper will be described in Section 3. Section 4 shows training details that include data preprocessing, training parameters, and postprocessing image stitching. Section 5 discusses the experimental situation and analyzes the results. The last section shows the conclusions of this paper.

## 2. Related work

### 2.1 Semantic segmentation for remote sensing imagery

The nature of semantic segmentation on remote sensing images is to classify houses, vehicles, roads, vegetation, oceanic ice and more at pixelwise precision. The remote sensing images come from aerial imagery or spectroscopy sensors. Early studies were primarily based on graph theory (Boykov, 2001; Di et al., 2017; Grady, 2006; Grady and Schwartz, 2004; Shi and Malik, 2000) using unsupervised learning. In 2009, Ye and Wang (Ye and Wang, 2009) proposed a new segmentation method on remote sensing imagery by combining the minimum spanning tree algorithm with Mumford Shah theory. Cui and Zhang (Cui and Zhang, 2011) proposed a multiscale and multilevel segmentation method that was also based on the minimum spanning tree in 2011, which performed well in very-high-resolution remote sensing image segmentation. For supervised learning, most segmentation methods learned with features that were selected by hand (Ouma et al., 2008; Reis and Taşdemir, 2011; Wang et al., 2016; Yu et al., 2016). These features are usually complex and can only express low- or mid-level descriptions.

With the development of remote sensing technologies, a large number of VHR remote sensing images can be easily obtained. These images have a wealth of contextual information, which make most of the traditional segmentation methods unsuitable. CNNs initially learn semantic representation of a pixel through patch-based training. The Fully Convolutional Network (FCN), which first came from Long et al. at Berkeley, is a milestone. It replaces all fully connected layers with convolutional neurons to allow arbitrary image size segmentation. Dilated convolution (Noh et al., 2015b) provides a larger receptive field under the same computational conditions. It could be used as a pooling operation to achieve dimensionality reduction. A large amount work has resulted in proposals to improve the dilated convolution, including atrous spatial pyramid pooling (Chen et al., 2017), fully connected CRF (Chen et al., 2018) and more. Inspirited by the encoder-decoder architecture, SegNet, DeconvNet, RefineNet (Lin et al., 2017) and U-Net were proposed to handle pixel-level segmentation on VHR remote sensing images. The gated segmentation network, GSN (Wang et al., 2017), recently achieved a competitive overall accuracy on the ISPRS 2D Potsdam dataset by adding gated thresholds to the short connections in DeconvNet.

## 2.2 Adaptive Networks in multiclass labeling

In early research, adaptive neural networks have helped to classify images through relaxed hierarchy structures in which a subset of confusing classes are allowed to be ignored (Deng and Satheesh, 2011; Gao and Koller, 2011; Griffin and Perona, 2008). The main difference between various existing methods is the way that the hierarchy is built. One common fundamental problem for these methods is that the assumption about the good separability of a binary partition of classes at each node is not valid when there is a large number of classes (Marszałek and Schmid, 2008). Thus, these methods do not scale well in terms of classification accuracy. Since deep learning has become a hot research topic, many CNN-based adaptive network methods have been proposed. Srivastava and Salakhutdinov (Srivastava and Salakhutdinov, 2013) introduced a category hierarchy in CNN-based methods for the first time. In these tree-like hierarchical CNN models, the upper nodes use the extracted common features to classify the images into superclasses, while the deeper nodes address finer features and perform further discrimination. Hierarchical Deep Convolutional Neural Networks, called HD-CNN (Yan et al., 2015), use common features, shared between images, to build a hierarchical CNN model for visual recognition.

The tree structures above are somewhat similar to our proposed tree structure. Differently, these networks are built before training, and the structure stays invariable in the process. Some dynamic methods have also been explored. Adaptive Neural Trees (Tanno et al., 2018) adaptively grow an architecture from primitive modules through adaptive neural trees (ANTs). Xiao et al. (Xiao et al., 2014) introduced a method that grows a tree-shaped network to accommodate new classes. Lifelong learning (Yoon et al., 2017) requires that a model adapts to new tasks while retaining its performance on previous tasks.

Some methods that can inherit previous training knowledge are explored. 'Learning without Forgetting' (Li and Hoiem, 2016) is such a method; it uses only new task data to train the network while preserving the original capabilities. Progressive Neural Networks (Rusu et al., 2016) learn to solve complex sequences of tasks by leveraging prior knowledge with lateral connections. Researchers have also designed spatially adaptive networks (Bengio et al., 2015; Figurnov et al., 2017) in which nodes in a layer are selectively activated. Others have developed cascade approaches (Leroux et al., 2017; Odena et al., 2017) that allow early exits based on confidence feedback.

Our proposed method constructs a tree-CNN block based on knowledge learned from progressive training. It does not predefine any structures and can be constructed automatically. The confusion matrix used to construct the tree is considered to be the recording memory. To the best of our knowledge, our work is the first to exploit both adaptive hierarchies and a deep neural network in a unified deep learning structure. The adaptive neural tree (Tanno et al., 2018) has also been proposed and involves a similar unified deep learning structure that combines the decision tree and CNN. Their work became available five months later than ours and does not solve semantic segmentation on remote sensing images.

## 3. Proposed method

For multiclass labeling tasks on remote sensing images, it is difficult to learn semantic representations for pixels because the easily confused classes are usually adjacent or interlaced in distribution. Based on the observation that classes can share some features while possessing their own properties, we attempt to address the multiclass labeling problem by hierarchically organizing the neural modules into a collection of subgroups and providing the easily confused classes more convolutional computations. To accomplish this approach in an end-to-end fashion, we propose a novel CNN architecture called TreeSegNet for semantic segmentation tasks on remote sensing images. The main framework is illustrated in Fig. 1.

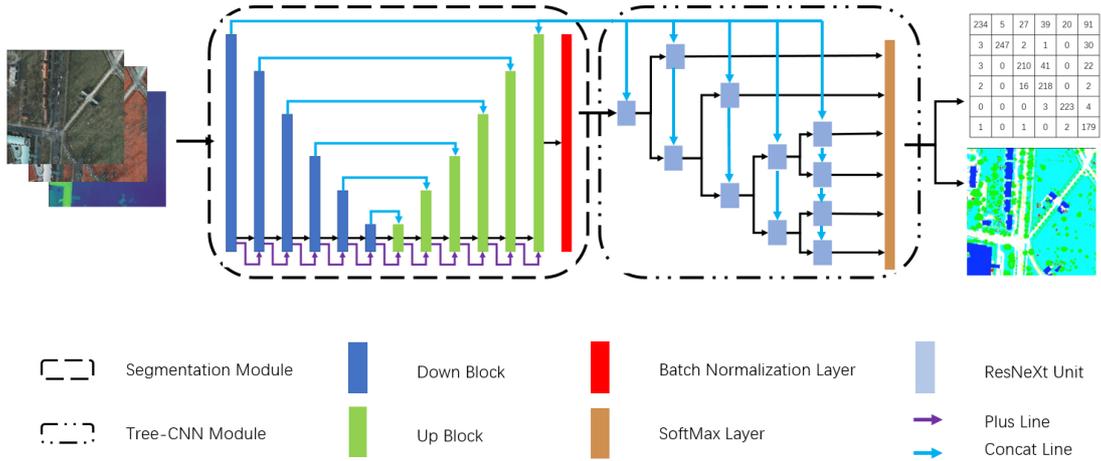

Fig. 1: the TreeSegNet framework

The TreeSegNet architecture is mainly composed of three parts: the segmentation module, the Tree-CNN block, and the concatenating connections. We choose the DeepUNet structure for the segmentation module. A batch normalization connects the DeepUNet module and the Tree-CNN block. The Tree-CNN block is built by an automatic construction method according to the confusion degree among the classes. For the first time, the method must train an initial model without the Tree-CNN block and compute the confusion matrix from the first-time segmentation results. In the subsequent passes, the Tree-CNN block is added and updated after each iteration. Specifically, after each iteration's training is completed, there will be a new confusion matrix calculated according to the segmentation results. The iterations are repeated until the structure of the Tree-CNN block no longer changes.

## 3.1 Segmentation module

As shown in Fig. 1, we use DeepUNet as the infrastructure network for pixelwise semantic segmentation. DeepUNet is based on VGG16 (Simonyan and Zisserman, 2014). It has two processing paths, as in U-Net. They are the contracting path and the expanding path. In the contracting path, the DownBlock is used as the basic feature extractor. It contains two convolutional layers and one pooling layer. Symmetrically, in the expanding path, the UpBlock is used as the upsampling block. Features are passed from the DownBlock to the UpBlock of the same level and then concatenated to perform convolution and upsampling. Both the DownBlocks and the UpBlocks use the residual operations and skip connections, which allows DeepUNet to be superior to other similar CNN architectures, such as U-Net and DeconvNet.

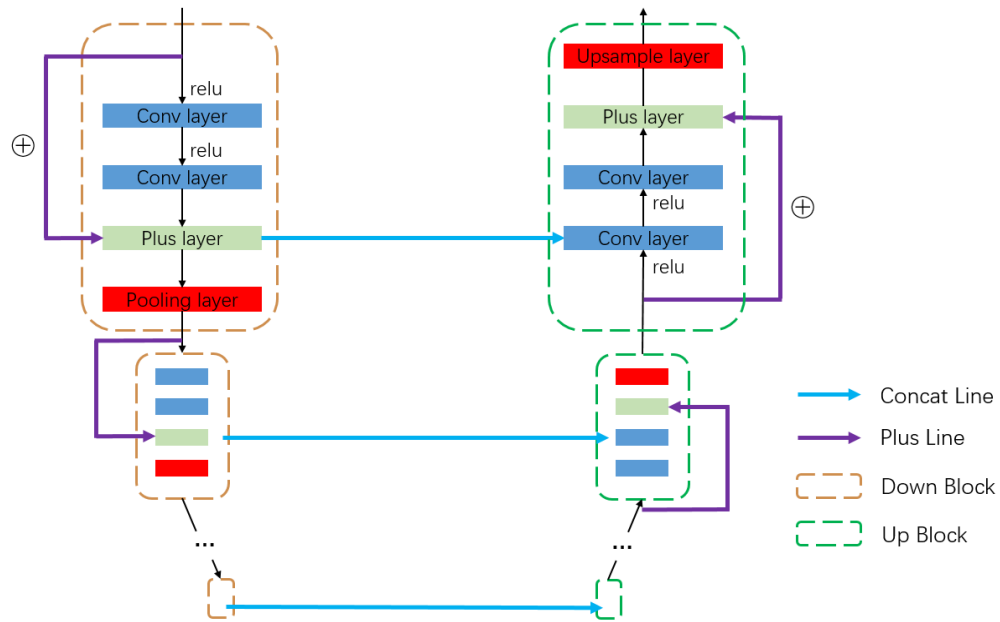

Fig. 2: DeepUNet network structure

Figure 2 shows the segmentation module of the DeepUNet. There are two types of connections. The blue lines are concatenating connections that transmit features from the left side to the right side. The purple lines are skip connections for performing residual operations. The DownBlocks are surrounded by the brown dotted rectangles, and the UpBlocks are surrounded by the green rectangles. In the DownBlock, two successive $3 \times 3$ convolutional layers are used instead of one $5 \times 5$ layer. This approach allows the block to obtain a receptive field of the same size but with fewer hyper-parameters. A Plus layer is added after the convolution operations and places the intermediate result into the pooling layer. The structure of the UpBlock is symmetric. In contrast, the pooling layer in the DownBlock is replaced by the upsampling layer in the UpBlock.

## 3.2 Tree-CNN block

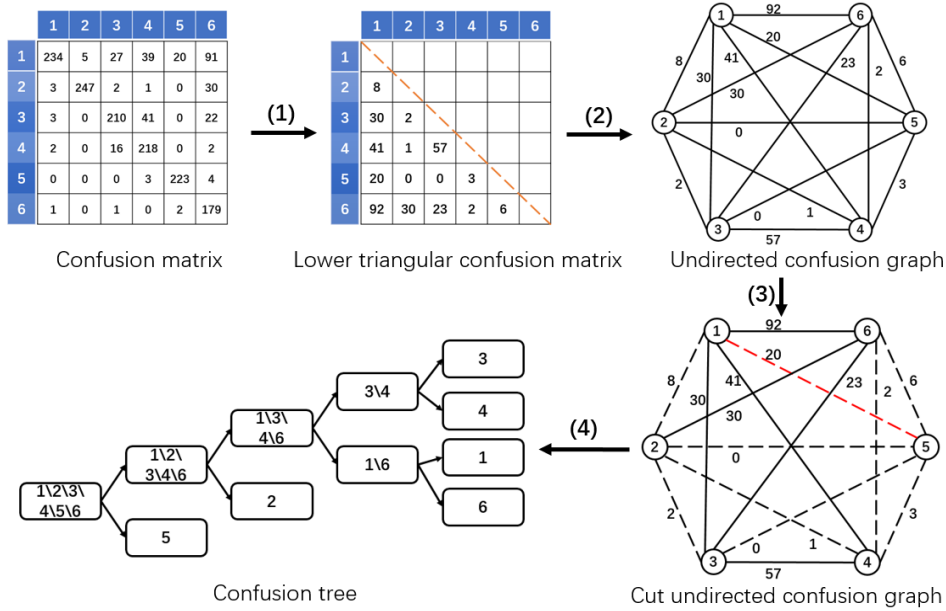

Fig. 3: The construction processing of Tree-CNN block

As shown in Fig. 3, the construction of Tree-CNN block is started by calculating the confusion matrix according to the first-time segmentation results. The confusion matrix, also known as an error matrix, is a specific table layout that allows visualization of the performance of an algorithm. Each row of the matrix represents the instances of a predicted class, while each column represents the instances of an actual class (or vice versa). With the confusion matrix, the following four steps are used to construct the Tree-CNN block:

(1) Calculating the lower triangular matrix;

(2) Building the undirected graph;

(3) Iterated the TreeCutting operation;

(4) Constructing the tree structure;

In confusion matrix A, we use $a_{ij}$ to represent the element of row i, column j; and $b_{ij}$ represents the element in the lower triangular confusion matrix B. According to the following formula, the lower triangular matrix B can be obtained by adding the values of the symmetric positions of the matrix A. Here, $b_{ij}$ represents the confusion degree between class i and class j.

$$b_{ij} = \begin{cases} a_{ij} + a_{ji}, & i > j \\ 0, & \text{otherwise} \end{cases}$$

The lower triangular matrix can be seen as an adjacency matrix that is associated with an undirected graph. In the graph, the nodes represent the indices of the classes, and the values of the connections represent the weights of the confusions. The numbers 1-6 represent imp_surf, building, low_veg, tree, car, and clutter, respectively.

### 3.2.1 Iterated TreeCutting

The undirected graph is then transformed into a binary tree structure by the proposed TreeCutting algorithm. The pseudocode of the algorithm is illustrated in Algorithm 1. The TreeCutting algorithm is executed iteratively. For each time, an edge with the minimum weight is always identified and is removed. When the current graph is divided into two subgraphs, the point sets of two subgraphs are added as the child nodes of the current root node. Each subgraph performs the same operation until no edges in the subgraph remain.

---

**Algorithm 1** The algorithm for the proposed **TreeCutting** operation

---

**Input:** The undirected confusion graph, $G(V, E)$

       The Point set of graph $G$, $V$

       The Edge set of graph $G$, $E$

**Output:** The binary tree-like structure $T$

1:    **while** $E$ still has edges $(E \neq \emptyset)$ **do**

2:        select $e \in E$ with minimum weight

3:        $G \leftarrow G(V, E\text{-}e)$

4:        **if** $G$ is still a complete graph **then**

5:            **continue**

6:        **elif** $G$ is cut into two subgraphs $G_1(V_1, E_1)$ & $G_2(V_2, E_2)$ **then**

7:            $T.leftchild \leftarrow V_1$

8:            $T.rightchild \leftarrow V_2$

9:            **TreeCutting**$(G_1, V_1, E_1)$

10:          **TreeCutting**$(G_2, V_2, E_2)$

11:     **end if**

12:  **end while**

---

### 3.2.2 Tree-CNN block

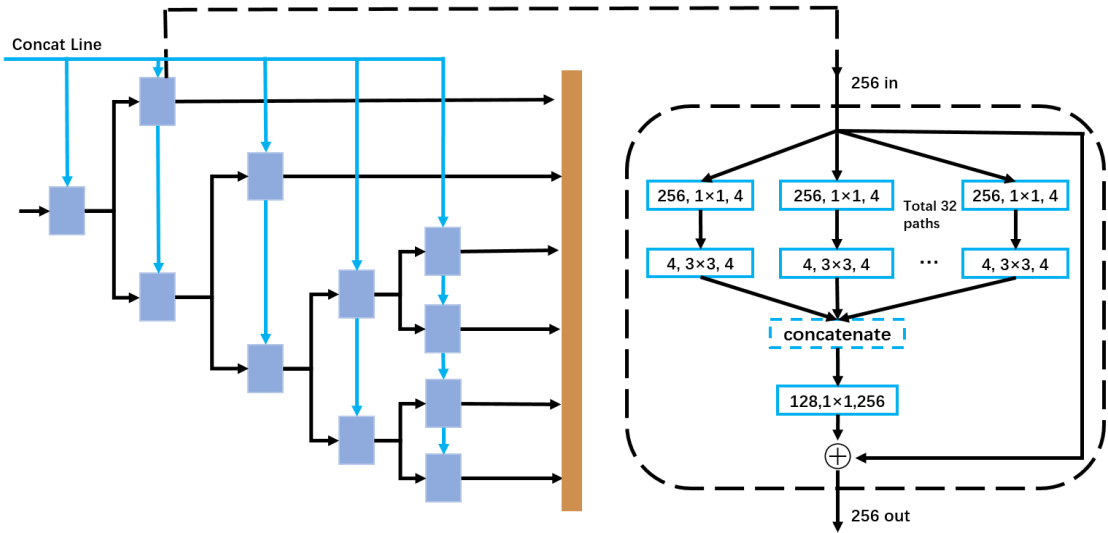

Fig. 4: The Tree-CNN block

The Tree-CNN block is the core of TreeSegNet, which can enhance the result of the previous semantic segmentation. The ISPRS dataset has six categories. Thus, the Tree-CNN block is illustrated with a binary tree that has six leaf nodes in Fig. 4. Each node is a ResNeXt unit shown on the right-hand side in Fig. 4. Through continuous optimization of back propagation, features of the distinct classes tend to go through the shorter path with fewer convolutional layers. In other words, the most easily confusing classes tend to choose the path that contains more neural layers for further feature extraction. For the Tree-CNN block, it is important to use the ResNeXt units. These units can not only avoid the gradient vanishing problem caused by deeper neural layers but also save graphics memories by reducing the number of hyper-parameters. The features that enter the ResNeXt unit have two sources: the output of the previous ResNeXt unit and the concatenating features.

Concatenating connections, a type of short connections, are used to transport features. The features output by the segmentation module are limited in size for the subsequent training of the Tree-CNN block. Thus, we connect the feature maps' output by the first convolutional layer to the input channels of the ResNeXt unit, as shown by the blue lines in Fig. 4. This type of connection passes the same feature map to all ResNeXt units of the Tree-CNN block, reusing the information extracted before. The number of features in the concatenating connections is relevant to the overall accuracy. In the experiment, we tested 16, 32 and 64 features to determine the best degree of feature reuse.

TreeSegNet is a fully convolutional neural network. It has no fully connected layers. After the Tree-CNN block, all of the features are passed to a $1 \times 1$ convolutional layer, and the weights are updated by the SOFTMAX loss function before the output is produced.

## 4. Implementation details

### 4.1 Data preprocessing

The Potsdam dataset of the ISPRS contains 2D multispectral remote sensing images. They are stored in the form of RGB, IRRG, and DSM. Before entering the TreeSegNet, the grouped channels of images must be split and transformed into five channels in a single image pattern.

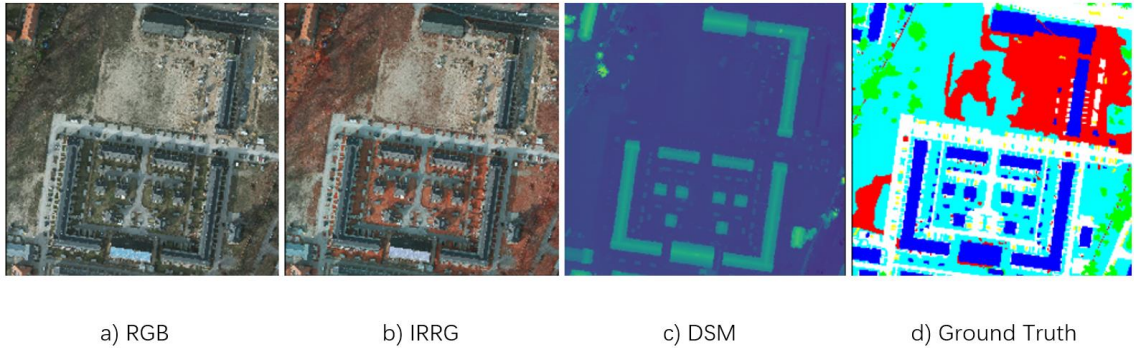

a) RGB    b) IRRG    c) DSM    d) Ground Truth

Fig. 5: An example of the Potsdam training images

Next, data enhancement is performed on the images by transformation, clipping, and rotation. Since the aerial imageries are orthogonal images that were photographed from the topside, they are rotated by 10 degrees repeatedly for 360 degrees to maximize the number of training data images. For each $6000 \times 6000$ pixel VHR image, 36 images are obtained after the rotation. According to the following steps, we calculate the maximum horizontal square in the rotated images shown in Fig. 6 b) and clip it out as in Fig. 6 c). All three squares share the same center positions. All of the newly expanded training images are clipped as the largest square after rotating the original image.

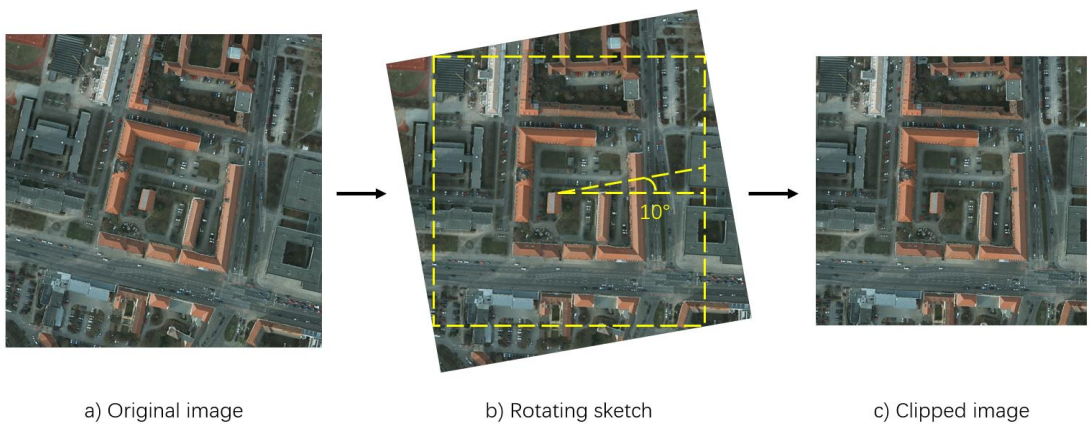

a) Original image    b) Rotating sketch    c) Clipped image

Fig. 6: An instance of image enhancement

## 4.2 Overlap tiles

The obtained square images are still too large for training. We must further clip the large images into tiles and test the tiles one by one from top left to bottom right with a sliding window approach. For very-high-resolution images, we propose an overlapping tiles strategy. Overlapping tiles are used not only because of the limitation of GPU memory but also because of the higher accuracy in the segmentation results.

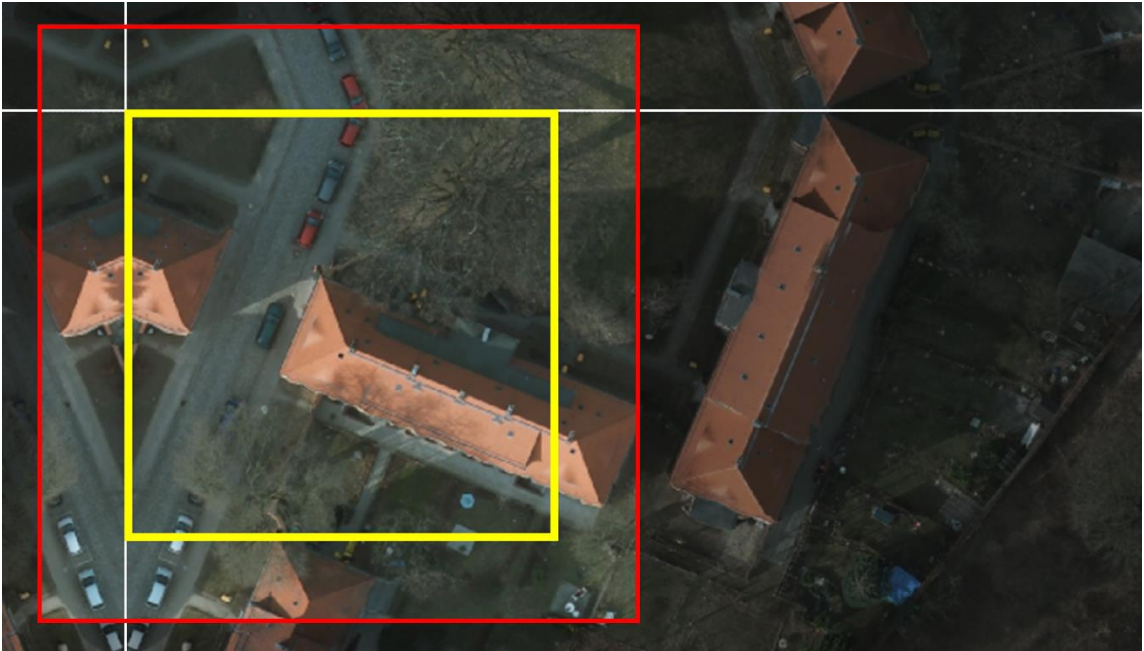

Fig. 7: Schematic diagram of overlapping tiles

The pixels surrounded by yellow lines in Fig. 7 indicate the valid area that must be predicted. And the pixels between the red lines and the yellow lines are called overlapping tiles or boundary fields; they provide context information to obtain a more accurate prediction. To predict the pixels in the border region of the image, the missing context is extrapolated by mirroring the input image. For the overlapping tiles, TreeSegNet computes the weight by a two-dimensional Gaussian function. σ is set to 0.5 by default. Pixels that are closer to the center have greater weight for subsequent stitching. Through a weighted summary, we composite the overlapping tiles and seamlessly stitch the whole segmented image.

$$g(x, y) = \frac{1}{2\pi\sigma^2} e^{-\left[\frac{(x-x_o)^2 + (y-y_o)^2}{2\sigma^2}\right]}$$

### 4.3 Training

TreeSegNet is implemented on MXNet. MXNet is a GPU friendly deep learning framework that provides two ways to program: the shallow embedded mode and the deep embedded mode. We use the deep embedded mode to implement TreeSegNet.

To minimize the overhead and make maximum use of the GPU memory, we favor large input tiles over a large batch size. For two NVidia 1080Ti GPUs, we choose $640 \times 640$ as the dimension of the input tiles, and hence, the batch size cannot be larger than 8. The epoch is initially set to 80. We use a high momentum (0.9). For the learning rate, the initial setting is 0.01. When the training is halfway done, the learning rate is adjusted to 0.001. When the number of training steps reaches 3/4 of the total, the learning rate is adjusted again to 0.0001.

During the training process, the Segmentation module and the Tree-CNN block are connected and trained as one unified network structure. In backpropagation, training errors are passed from the Tree-CNN block to the Segmentation module. It should be noted that the structure of the Tree-CNN block will update as the confusion matrix changes throughout the training process. However, at every iteration of the process, the structure of the Tree-CNN block remains invariable.

## 5. Experiments and analysis

Experiments are performed on the ISPRS Potsdam 2D labeling dataset to validate the effectiveness of the TreeSegNet. We first compare the proposed method with classical methods, including DST_5 (Sherrah, 2016), RIT_7 (Liu et al., 2017) and others. We also compare our method with DeepUNet on the same Potsdam 2D remote sensing images under the same experimental environments. Then, detailed analyses are performed by adjusting the training parameters, tuning the network structures, and using different Tree-CNN structures.

## 5.1. Experimental environments

The experiments are conducted on a laboratory computer. Its configuration is shown in Table 1. The operating system has Ubuntu 16.04 installed. The main required packages include python 2.7, CUDA8.0, cuDNN7, Tensorflow1.1.0, Caffe, Keras1.2.0, MXNet0.10.0 and more.

**Table 1.** Experimental environment

| | |
|---|---|
| CPU | Intel (R) Core (TM) i7-4790K 4.00 Hz |
| GPU | GeForce GTX1080 Ti |
| RAM | 20 GB |
| Hard disk | Toshiba SSD 512 G |
| System | Ubuntu 16.04 |

## 5.2 Dataset

The experiments use the 2D semantic labeling contest Potsdam dataset released by ISPRS Commission II/4, which is a remote sensing research dataset that describes the environment and surroundings in and around the city. This dataset, using a digital aerial camera to take a vertical shot of multiple parts of a site, includes 38 image patches, and each consists of a $6000 \times 6000$ resolution true orthophoto (TOP) and its corresponding digital surface model (DSM). In all 38 image patches, labeled ground truth is provided for only 24 patches for network training. The ground truth of the remaining 14 patches will remain unreleased, to be used for the evaluation of submitted results. It includes the six most common land cover classes, including impervious surfaces (imp_surf), buildings (building), low vegetation (low_veg), trees (tree), cars (car), and clutter/background (clutter). The dataset contains city and suburban artificial surfaces that make extensive use, of filling mixture formations (such as concrete, steel, and wooden roofs) and seminatural environments (such as artificial grassland and bare soil). ISPRS shows that the suburbs of the city contain a small number of trees, bushes, cars, and sundries.

### 5.3 Evaluation matrix

The performance of TreeSegNet is evaluated on the ISPRS Potsdam 2D dataset for overall accuracy (OA) and F1 score on each class. The OA measures the global accuracy of the semantic segmentation, which provides information about the rate of correctly classified pixels. The OA and F1 score can be calculated by the following formulas:

$$OA = \frac{tp + tn}{p + n}$$

$$F1 = 2 \times \frac{precision \times recall}{precision + recall}$$

$$precision = \frac{tp}{tp+fp} \qquad recall = \frac{tp}{tp+fn}$$

In the confusion matrix, the True Positive (TP) is the value of the corresponding diagonal elements. The False Positive (FP) is computed from the summation of the column, while the False Negative (FN) is the summation of the row, excluding the main diagonal element.

### 5.4 Results and analysis

In Section 5.4.1 and 5.4.2, when TreeSegNet compares the experimental results with most state-of-the-art methods, we use all 24 labeled images for training and 14 unlabeled images for testing. In Section 5.4.3 and 5.4.4, to analyze the structure and parameters of TreeSegNet, we divided the 24 labeled images into a training set (18 images) and a validation set; the validation set contained the remaining 5 images (image numbers 7_7, 7_8, 7_9, 7_11, 7_12). The image numbered 7_10 has error annotations, which will bring about the degradation of the network segmentation performance. It was removed from further experiments.

5.4.1 Overall performance

On the ISPRS Potsdam website, we have not found public results by U-Net and DeconvNet that obtain great success for semantic segmentation. To verify the proposed method, we reimplemented U-Net (BUCT1) and DeconvNet, and applied them to Potsdam 2D semantic labeling.

We submitted our results on the unlabeled test images to the ISPRS organizers for evaluation. The TreeSegNet (BUCTY4) ranks 1st both in mean F1 score and OA, compared with other published works. The result images are visually illustrated in Fig. 8, and the detailed numerical scores are listed in Table 2. The methods shown include SVL_1, DST_5 (Sherrah, 2016), UZ_1 (Volpi and Tuia, 2017), RIT_L7 (Liu et al., 2017), KLab_2 (Kemker et al., 2018), GSN (Wang et al., 2017), DeconvNet (Noh et al., 2015a), U-Net (Ronneberger et al., 2015), DeepUNet (Li et al., 2018), and TreeSegNet. As seen from Table 2, the segmentation result of TreeSegNet takes first place in each category, except for the tree class.

In comparison with DST_5 which ranked second in terms of OA, TreeSegNet obtained 0.9% and 1.1% higher F1 scores in the building and car categories. The performance of U-Net is also competitive. Compared with U-Net, TreeSegNet obtained 0.5%, 0.3%, 0.5%, 0.9% and 1.9% higher F1 scores in the imp_surf, building, low_veg, tree and car classes.

**Table 2**. Quantitative comparisons between our method and other related methods

(already published) on the ISPRS test set.

| Method | Imp_surf | Building | Low_veg | Tree | Car | OA | Mean F1 |
|--------|----------|----------|---------|------|-----|-----|---------|
| SVL_1 | 83.5 | 91.7 | 72.2 | 63.2 | 62.2 | 77.8 | 74.6 |
| DST_5 | 92.5 | 96.4 | 86.7 | 88.0 | 94.7 | 90.3 | 91.7 |
| UZ_1 | 89.3 | 95.4 | 81.8 | 80.5 | 86.5 | 85.8 | 86.7 |
| RIT_L7 | 91.2 | 94.6 | 85.1 | 85.1 | 92.8 | 88.4 | 89.8 |
| KLab_2 | 89.7 | 92.7 | 83.7 | 84.0 | 92.1 | 86.7 | 88.4 |
| GSN | 92.2 | 95.1 | 83.7 | **89.9** | 82.4 | 90.3 | 88.7 |
| DeconvNet | 82.4 | 85.8 | 69.1 | 66.7 | 75.3 | 76.3 | 75.8 |
| U-Net | 92.4 | 97.0 | 86.3 | 86.4 | 93.9 | 90.0 | 91.1 |
| DeepUNet | 92.8 | 97.4 | 84.5 | 85.5 | 95.1 | 89.4 | 91.1 |
| TreeSegNet | **92.9** | **97.3** | **86.8** | 87.3 | **95.8** | **90.5** | **92.0** |

In Fig. 8, we mainly show the visual results of "top_Potsdam_5_13_class.tif". The figures are divided into two groups. The first group includes the first two columns in Fig. 8, while the second group consists of the last two columns. The first column of each group is the original test image and its segmentation results from related methods. The second column displays the evaluation images, where red pixels represent those that are wrongly segmented and green pixels vice versa. Since the GSN did not disclose the results on the Potsdam dataset, the detailed segmentation images of the GSN is not included in Fig. 8. The first group shows the overall segmented images. It can be found that TreeSegNet produces fewer segmentation errors, especially in the tree and low_veg categories. The second group magnifies the local details. The TreeSegNet accurately segmented the trees without any adhesion of pixels. Most other methods segmented the trees into a blurred area. However, each tree is independent of each other and does not stick to each other. In addition, most other methods mistook the building's corner for the clutter category in the lower-left corner of the second group, while TreeSegNet got the correct segmentation result.

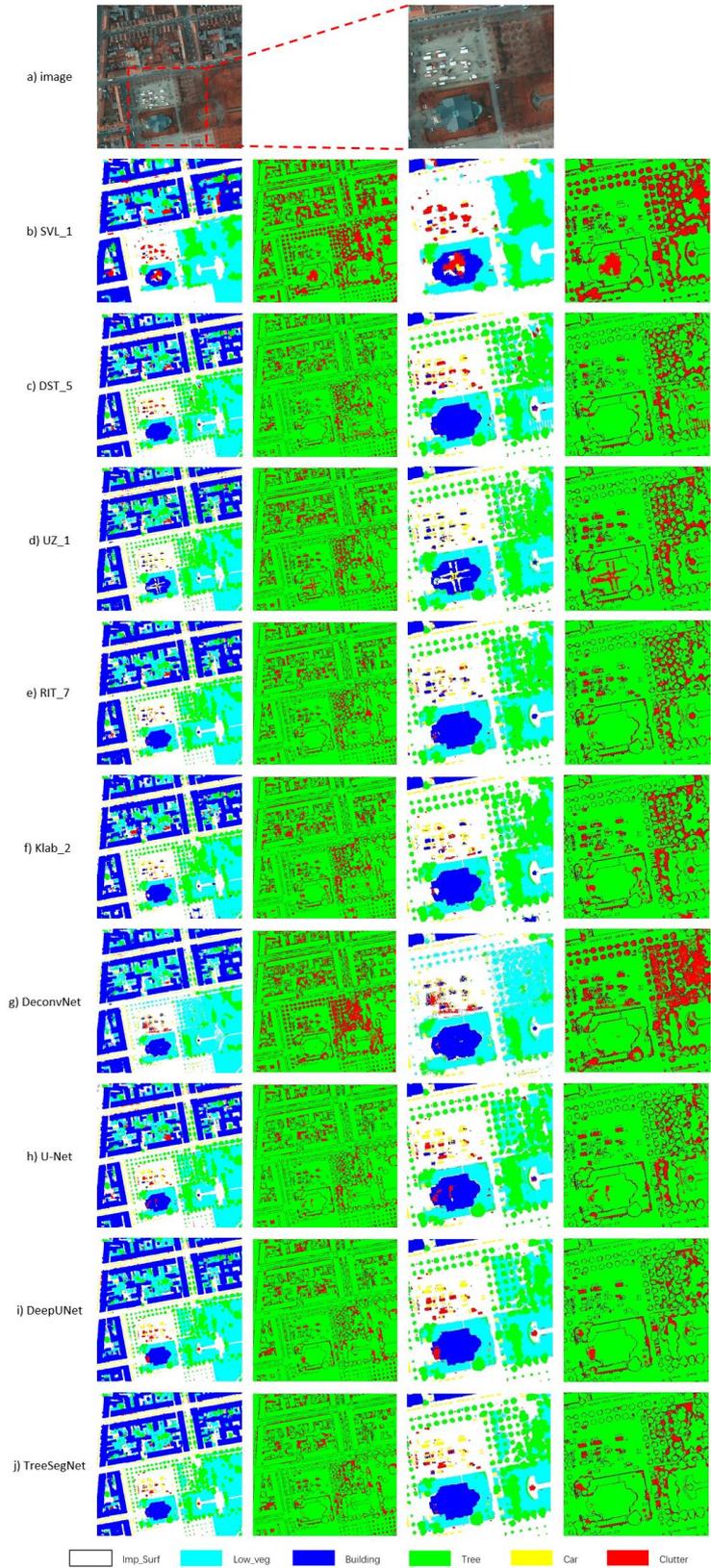

Fig. 8: Overall and local detailed results of "top_Potsdam_5_13_class.tif". Images come from the website of ISPRS 2D Semantic Labeling Contest.

It should be noted that the results of the above-listed methods are not to be exactly compared under the same conditions. For example, KLab_2 trains the model on its own expanded dataset, named RIT-18. Without training on the extended data, it may lead to a degradation on the segmentation results for their network. Different but not contradictory, the TreeSegNet aims to establish an end-to-end network architecture for addressing easily confused categories, suitable for VHR images.

5.4.2 Improvement by TreeSegNet

We adopted U-Net, DeepUNet, and TreeSegNet for comparisons to show the superiority of Tree-CNN block. Table 3 below shows the detailed numerical scores of U-Net, DeepUNet, and TreeSegNet.

**Table 3.** Segmentation results for DeepUNet and TreeSegNet.

| Method | Imp_surf | Building | Low_veg | Tree | Car | OA | Mean_F1 |
|--------|----------|----------|---------|------|-----|-----|---------|
| U-Net | 92.4 | 97.0 | 86.3 | 86.4 | 93.9 | 90.0 | 91.1 |
| DeepUNet | 92.8 | **97.4** | 84.5 | 85.5 | 95.1 | 89.4 | 91.1 |
| TreeSegNet | **92.9** | 97.3 | **86.8** | **87.3** | **95.8** | **90.5** | **92.0** |

The training of DeepUNet is the first phase of TreeSegNet. It provided the first pass running result for TreeSegNet. With the Tree-CNN block, TreeSegNet improved significantly. The OA increased from 89.4% to 90.5%, and the mean F1 score increased from 91.1% to 92.0%. Both evaluation indicators reached the highest values.

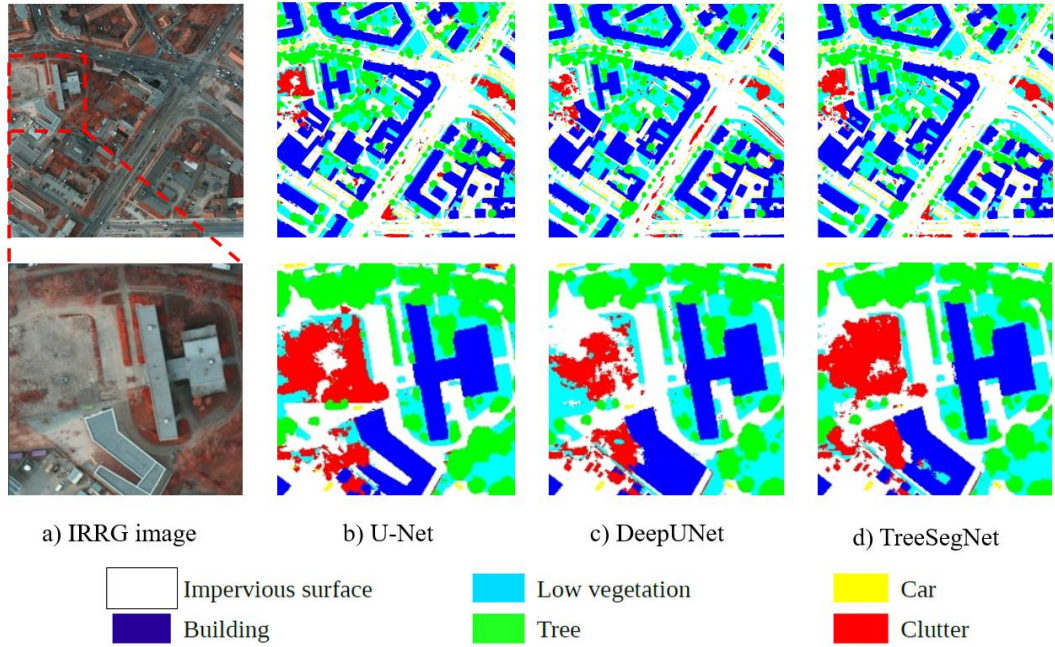

a) IRRG image      b) U-Net      c) DeepUNet      d) TreeSegNet

| | | |
|---|---|---|
| ☐ Impervious surface | ☐ Low vegetation | ☐ Car |
| ■ Building | ■ Tree | ■ Clutter |

Fig. 9: Global and local details of segmentation on "top_Potsdam_5_15_class.tif"

Figure 9 illustrates the original image and the segmentation sample of "top_Potsdam_5_15_class.tif". TreeSegNet ranked first in the remaining four categories, including building, low_veg, car, and tree. We have defined the two easily confused category pairs: 1) imp_surf and clutter, and 2) low_veg and tree. With regard to the categories of low_veg and trees, the F1 scores of TreeSegNet increased by 2.2% and 1.3% compared with the original DeepUNet. With regard to another pair of easily confused categories, imp_surf and clutter, TreeSegNet is 0.1% and 6.4% higher than DeepUNet, respectively. TreeSegNet greatly improved the segmentation accuracy of clutter category. In the second row of Fig. 9, the boundary segmented by TreeSegNet between the clutter and the building categories was accurate and smooth.

**Table 4**. The evaluation scores of "top_Potsdam_5_15_class.tif"

| Method | Imp_surf | Building | Low_veg | Tree | Car | Clutter | OA |
|--------|----------|----------|---------|------|-----|---------|-----|
| U-Net | **93.6** | 97.3 | 81.3 | 85.0 | 91.6 | **62.0** | 90.1 |
| DeepUNet | 92.2 | 97.3 | 80.8 | 86.7 | 95.7 | 47.8 | 89.0 |
| TreeSegNet | 93.3 | **97.6** | **83.0** | **88.0** | **96.0** | 59.7 | **90.6** |

Figure 10 shows the segmentation results of "top_Potsdam_6_14_class.tif". The corresponding accuracy of the six categories of U-Net, DeepUNet, and TreeSegNet are shown in Table 5. TreeSegNet obtains the highest F1 scores for all categories except for the building category. Compared with DeepUNet, TreeSegNet has increased F1 scores by 0.1% and 9.7% on the imp_surf and clutter categories, respectively. The increases in low_veg and tree are 1.6% and 0.4%. For tiny objects like cars, our results are still satisfactory by reducing the adhesion of pixels and the bubble-effect. In addition, Fig.10 clearly shows that the clutter area segmented by TreeSegNet has a smoother and more coherent boundary with fewer errors.

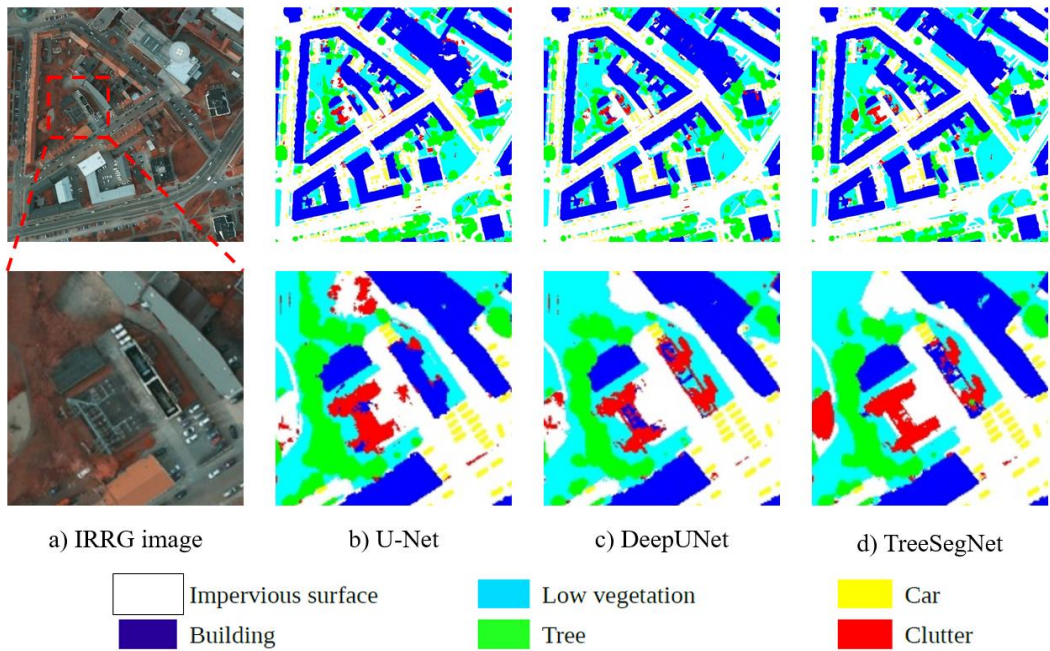

a) IRRG image      b) U-Net      c) DeepUNet      d) TreeSegNet

    Impervious surface      Low vegetation      Car

    Building      Tree      Clutter

Fig. 10: Global and local details of segmentation on "top_Potsdam_6_14_class.tif"

**Table 5**. The evaluation scores of "top_Potsdam_6_14_class.tif"

| Method | Imp_surf | Building | Low_veg | Tree | Car | Clutter | OA |
|---|---|---|---|---|---|---|---|
| U-Net | 95.2 | 97.1 | 88.0 | 86.3 | 96.8 | 38.4 | 93.0 |
| DeepUNet | 95.7 | **97.7** | 87.9 | 88.0 | 97.2 | 47.1 | 93.6 |
| TreeSegNet | **95.8** | 97.6 | **89.5** | **88.4** | **97.8** | **56.8** | **94.1** |

For the next groups of experiments in Section 5.4.3 and 5.4.4, we divide the labeled Potsdam datasets into two parts, 18 images as the training set, and 5 images (image numbers 7_7, 7_8, 7_9, 7_11, 7_12) as the validation set. Following results are reported on the validation set if not specified.

5.4.3 Different tree structures

To understand the benefit of the Tree-CNN structures, we perform a detailed analysis in Fig. 11 and Table 6. There are two usual ways to increase the complexity of the network structure and bring about better performance: to widen the network or to deepen it. The Tree-CNN structures have many choices, for example, using a binary balanced tree structure (widen the network) or directly increasing the convolutional layers by adding a 'straight tree' structure (deepen the network). Thus, it is necessary to prove that the higher OA of TreeSegNet is not brought about by having a redundant deeper or wider network structure. We designed two special tree structures: the elegant balance tree and the straight tree structure (shown in Fig. 11.a and Fig. 11.b), and we trained them under the same condition to further understand the TreeSegNet. In fact, the TreeSegNet is dynamically and adaptively constructed. For the first iteration, it is trained without the Tree-CNN block. Then, the Tree-CNN block is constructed from the confusion matrix derived from the initial segmentation results which is showed in Fig.11.c. Next, the Tree-CNN block is updated according the same procedure to Fig. 11.d. These steps are iterated until the structure of the Tree-CNN block no longer changes. According to the experimental results, the evolution of the confusion trees changes from none to Fig. 11.c, and then to Fig. 11.d. Table 6 shows the comparison results. TreeSegNet with the iterated confusion tree structure shown in Fig. 11.d obtained the highest OA of 90.66%. The number of features on the concatenating connection that TreeSegNet used for this set of experiments is 64.

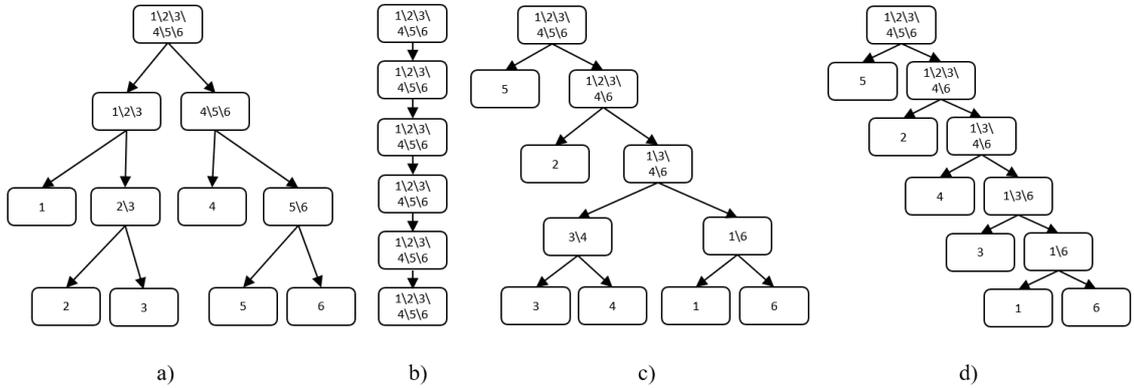

Fig. 11: Different tree structures in Table 6

**Table 6**. The OA scores of the validation set for different structures

| Structure | OA |
|---|---|
| TreeSegNet with no tree | 88.43 |
| TreeSegNet + balanced tree (a in Fig. 11) | 89.88 |
| TreeSegNet + straight structure (b in Fig. 11) | 89.22 |
| TreeSegNet + 1 pass (c in Fig. 11) | 90.61 |
| TreeSegNet + 2 passes (d in Fig. 11) | **90.66** |

Our experiments show that the Tree-CNN block adaptively learned through network segmentation results performs best on the segmentation results, rather than artificial designed ones. As the structure of the Tree-CNN block updates, the OA of the network segmentation results will be improved step by step.

5.4.4 Different concatenating features

The number of features on the concatenating connection is a key parameter. We tried out different numbers of features in the concatenating connections, using 16, 32, and 64. The OA for three different feature numbers is shown in Fig. 12. According to the highest OA of 90.6%, we selected 64 as the number of features in the concatenating connections. The tree structure used is similar to Fig. 11. d.

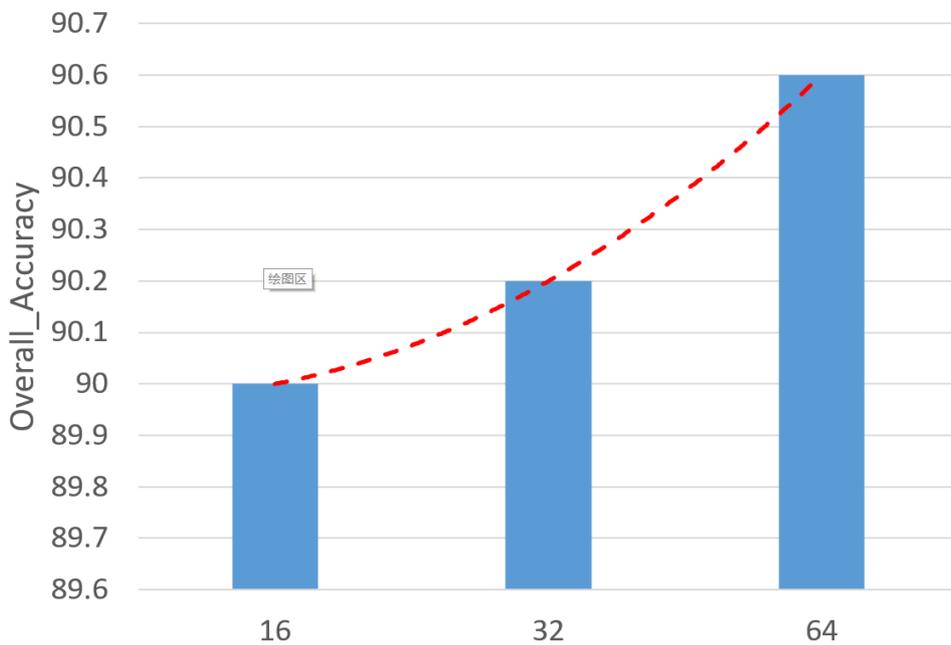

Fig. 12: Influence of the number of features in the concatenating connections on the OA.

## 6. Conclusions

In this paper, we propose a new approach to addressing the semantic segmentation of very-high-resolution remote sensing images, using TreeSegNet with an automatic constructed Tree-CNN block. In the experiments on the ISPRS Potsdam dataset, the F1 scores of the easily confused categories all improved. Finally, we obtained the highest OA among the already opened list of state-of-the-art methods.

In the future, other data augmentation methods could be considered, for example, using GAN. Since we use the original DSM for training and testing, feature fusion or normalization of DSM could also be added to the proposed method, which could further improve the accuracy.

## 7. Acknowledgments

The authors would like to acknowledge the provision of the datasets by ISPRS and BSF Swiss photo, which were released in conjunction with the ISPRS, led by ISPRS WG II/4.

High-Resolution Satellite Image Classification. Remote Sens. 8, 259.

https://doi.org/10.3390/rs8030259

# Appendices

## A. Further explanation on our proposed TreeCutting Operation

In this section, we expect to point out the gap between our TreeCutting operations and the minimal spanning trees and the Min-Cut/Max-Flow Algorithm.

A.1 Process instance of TreeCutting operation

In the beginning, we have a lower triangular matrix, which is calculated from the previous segmentation results. We provide an instance of the TreeCutting operation in Fig.14. From left to right, the sub-columns represent lower triangular matrixes, undirected graphs, and constructed dendrograms respectively. The blue cells in the left subfigures indicate that the correspondent edges have been cut off. In step II, for example, the edges with weights of 0, 1, 2, 3, and 6 have been cut off, and the edge with the weight of 20 will be the next one. The current cutting operation splits the graph into two subgraphs. The left child contains one node--number 5. The right child has 5 nodes which will be split out in a nested fashion. The detailed process is demonstrated step by step in Fig. 13.

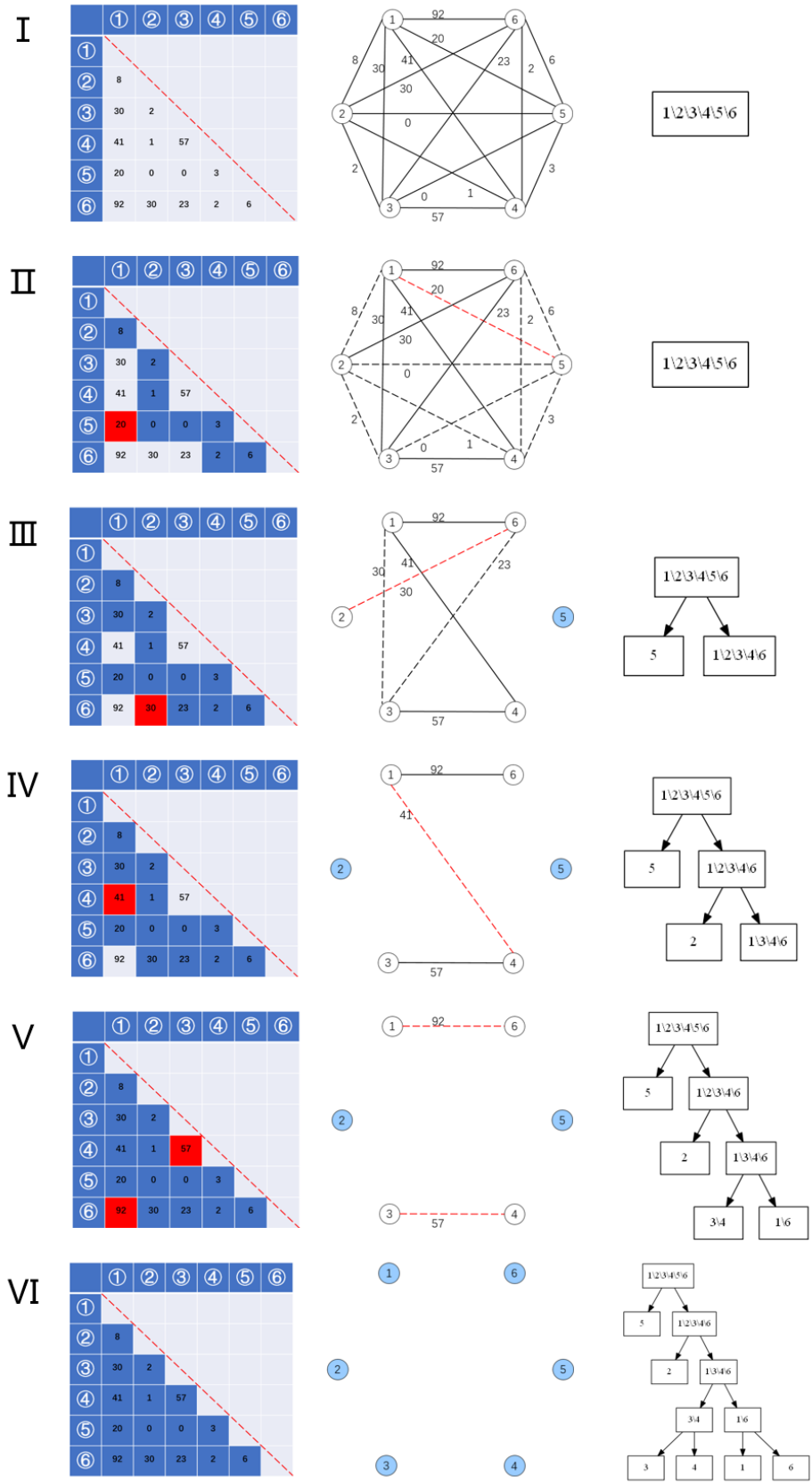

Fig. 13: Processing flow of TreeCutting algorithm

A.2 Comparison of related algorithms

Our TreeCutting algorithm has similarities to the minimal spanning trees(Wikipedia, 2015) and the Min-Cut/Max-Flow Algorithm (Boykov and Kolmogorov, 2004). In the following text, we only select some points that are sufficient to prove that our TreeCutting algorithm is different from the mentioned ones.

One of the most important conditions of the minimum spanning tree algorithm is to maintain the connectivity of the graph, that is, all nodes in the original graph are accessible to each other. Our TreeCutting algorithm usually cuts off one edge with minimum weight from the original graph and then divides it into two subgraphs. The connectivity of the original graph is broken. So, our TreeCutting algorithm is completely different from the minimum spanning trees.

For the Min-Cut/Max-Flow Algorithm, there is always a starting point and a terminal point, while the TreeCutting algorithm has no corresponding points. In addition, different from the Min-Cut/Max-Flow Algorithm, out TreeCutting algorithm requires that the edge with the minimum weight always been found and cut off. When the Min-Cut/Max-Flow Algorithm sometimes cuts the original graph into the subgraphs, the cut edges are not necessarily the collection of all the smallest ones. An example is given below:

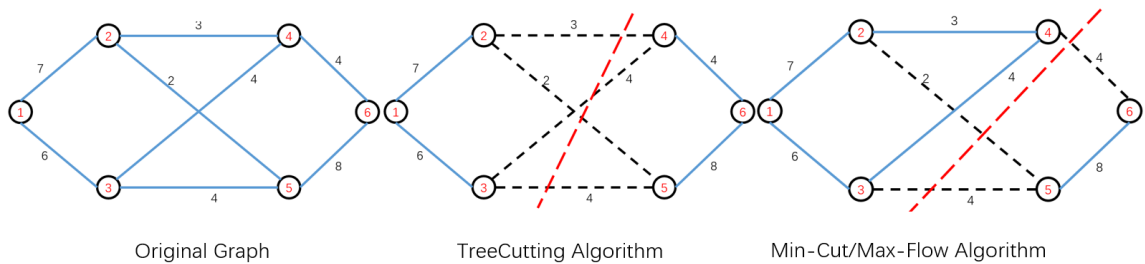

Fig. 14: An instance of TreeCutting and the Min-Cut/Max-Flow Algorithm

The one left is the original graph, the middle one is the result of our TreeCutting operation, and the right is the result of the Min-Cut/Max-Flow Algorithm (Any point in the set{1, 2, 3, 4} can be used as the starting point, and anyone in the set {5, 6} as the terminal point). Our proposed algorithm cuts the edge set {2, 3, 4, and 4}. But the Min-Cut/Max-Flow Algorithm cuts {2, 4, 4} where a smaller edge of 3 is not selected in.

## B. Detailed Descriptions for 2D Semantic Labeling Contest – Potsdam

We have taken some specific descriptions of the Potsdam dataset from the website that may explain some misunderstandings about the ground truth, spatial resolution, and patch numbering. The original text is as follows:

The data set of Potsdam contains 38 patches (of the same size), each consisting of a true orthophoto (TOP) extracted from a larger TOP mosaic, see Fig.17 below and a DSM.

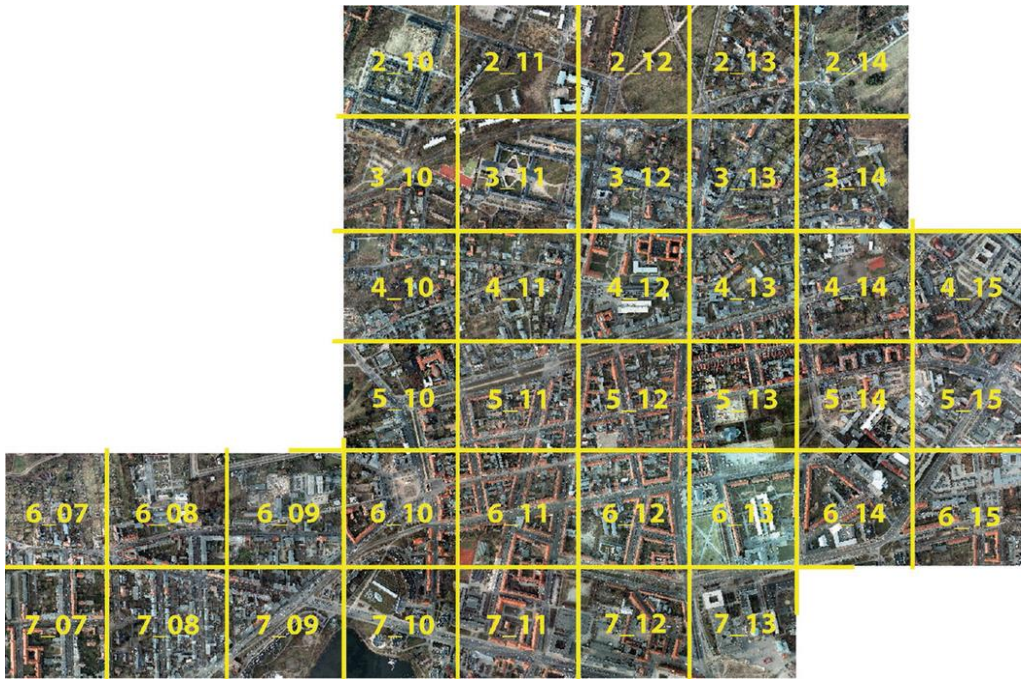

Figured. 15: The corresponding patches to the patch numbers in the dataset

Outlines of all patches given in Fig. 15 overlaid with the true orthophoto mosaic. Numbers refer to the individual patch numbers, encoded in the filenames as well.

The ground sampling distance of both, the TOP and the DSM, is 5 cm. The DSM was generated via dense image matching with Trimble INPHO 5.6 software and Trimble INPHO OrthoVista was used to generate the TOP mosaic. The TOP comes as TIFF files in different channel compositions, where each channel has a spectral resolution of 8bit:

- IRRG: 3 channels (IR-R-G)
- RGB: 3 channels (R-G-B)
- RGBIR: 4 channels (R-G-B-IR)

The DSM are TIFF files with one band; the grey levels (corresponding to the DSM heights) are encoded as 32-bit float values.

Labeled ground truth is provided for only one part of the data. The ground truth of the remaining scenes will remain unreleased and stays with the benchmark test organizers to be used for evaluation of submitted results. Participants shall use all data with ground truth for training (see Fig. 16) or internal evaluation of their method.

| TOP RGBIR | TOP IRRG | TOP RGB | DSM | GT |
|---|---|---|---|---|
| top_potsdam_2_10_RGBIR | top_potsdam_2_10_IRRG | top_potsdam_2_10_RGB | dsm_potsdam_02_10 | top_potsdam_2_10_label |
| top_potsdam_2_11_RGBIR | top_potsdam_2_11_IRRG | top_potsdam_2_11_RGB | dsm_potsdam_02_11 | top_potsdam_2_11_label |
| top_potsdam_2_12_RGBIR | top_potsdam_2_12_IRRG | top_potsdam_2_12_RGB | dsm_potsdam_02_12 | top_potsdam_2_12_label |
| top_potsdam_2_13_RGBIR | top_potsdam_2_13_IRRG | top_potsdam_2_13_RGB | dsm_potsdam_02_13 | |
| top_potsdam_2_14_RGBIR | top_potsdam_2_14_IRRG | top_potsdam_2_14_RGB | dsm_potsdam_02_14 | |
| top_potsdam_3_10_RGBIR | top_potsdam_3_10_IRRG | top_potsdam_3_10_RGB | dsm_potsdam_03_10 | top_potsdam_3_10_label |
| top_potsdam_3_11_RGBIR | top_potsdam_3_11_IRRG | top_potsdam_3_11_RGB | dsm_potsdam_03_11 | top_potsdam_3_11_label |
| top_potsdam_3_12_RGBIR | top_potsdam_3_12_IRRG | top_potsdam_3_12_RGB | dsm_potsdam_03_12 | top_potsdam_3_12_label |
| top_potsdam_3_13_RGBIR | top_potsdam_3_13_IRRG | top_potsdam_3_13_RGB | dsm_potsdam_03 13 | |
| top_potsdam_3_14_RGBIR | top_potsdam_3_14_IRRG | top_potsdam_3_14_RGB | dsm_potsdam_03_14 | |
| top_potsdam_4_10_RGBIR | top_potsdam_4_10_IRRG | top_potsdam_4_10_RGB | dsm_potsdam_04_10 | top_potsdam_4_10_label |
| top_potsdam_4_11_RGBIR | top_potsdam_4_11_IRRG | top_potsdam_4_11_RGB | dsm_potsdam_04_11 | top_potsdam_4_11_label |
| top_potsdam_4_12_RGBIR | top_potsdam_4_12_IRRG | top_potsdam_4_12_RGB | dsm_potsdam_04_12 | top_potsdam_4_12_label |
| top_potsdam_4_13_RGBIR | top_potsdam_4_13_IRRG | top_potsdam_4_13_RGB | dsm_potsdam_04_13 | |
| top_potsdam_4_14_RGBIR | top_potsdam_4_14_IRRG | top_potsdam_4_14_RGB | dsm_potsdam_04_14 | |
| top_potsdam_4_15_RGBIR | top_potsdam_4_15_IRRG | top_potsdam_4_15_RGB | dsm_potsdam_04_15 | |
| top_potsdam_5_10_RGBIR | top_potsdam_5_10_IRRG | top_potsdam_5_10_RGB | dsm_potsdam_05_10 | top_potsdam_5_10_label |
| top_potsdam_5_11_RGBIR | top_potsdam_5_11_IRRG | top_potsdam_5_11_RGB | dsm_potsdam_05_11 | top_potsdam_5_11_label |
| top_potsdam_5_12_RGBIR | top_potsdam_5_12_IRRG | top_potsdam_5_12_RGB | dsm_potsdam_05_12 | top_potsdam_5_12_label |
| top_potsdam_5_13_RGBIR | top_potsdam_5_13_IRRG | top_potsdam_5_13_RGB | dsm_potsdam_05_13 | |
| top_potsdam_5_14_RGBIR | top_potsdam_5_14_IRRG | top_potsdam_5_14_RGB | dsm_potsdam_05_14 | |
| top_potsdam_5_15_RGBIR | top_potsdam_5_15_IRRG | top_potsdam_5_15_RGB | dsm_potsdam_05_15 | |
| top_potsdam_6_7_RGBIR | top_potsdam_6_7_IRRG | top_potsdam_6_7_RGB | dsm_potsdam_06_07 | top_potsdam_6_7_label |
| top_potsdam_6_8_RGBIR | top_potsdam_6_8_IRRG | top_potsdam_6_8_RGB | dsm_potsdam_06_08 | top_potsdam_6_8_label |
| top_potsdam_6_9_RGBIR | top_potsdam_6_9_IRRG | top_potsdam_6_9_RGB | dsm_potsdam_06_09 | top_potsdam_6_9_label |
| top_potsdam_6_10_RGBIR | top_potsdam_6_10_IRRG | top_potsdam_6_10_RGB | dsm_potsdam_06_10 | top_potsdam_6_10_label |
| top_potsdam_6_11_RGBIR | top_potsdam_6_11_IRRG | top_potsdam_6_11_RGB | dsm_potsdam_06_11 | top_potsdam_6_11_label |
| top_potsdam_6_12_RGBIR | top_potsdam_6_12_IRRG | top_potsdam_6_12_RGB | dsm_potsdam_06_12 | top_potsdam_6_12_label |
| top_potsdam_6_13_RGBIR | top_potsdam_6_13_IRRG | top_potsdam_6_13_RGB | dsm_potsdam_06_13 | |
| top_potsdam_6_14_RGBIR | top_potsdam_6_14_IRRG | top_potsdam_6_14_RGB | dsm_potsdam_06_14 | |
| top_potsdam_6_15_RGBIR | top_potsdam_6_15_IRRG | top_potsdam_6_15_RGB | dsm_potsdam_06_15 | |
| top_potsdam_7_7_RGBIR | top_potsdam_7_7_IRRG | top_potsdam_7_7_RGB | dsm_potsdam_07_07 | top_potsdam_7_7_label |
| top_potsdam_7_8_RGBIR | top_potsdam_7_8_IRRG | top_potsdam_7_8_RGB | dsm_potsdam_07_08 | top_potsdam_7_8_label |
| top_potsdam_7_9_RGBIR | top_potsdam_7_9_IRRG | top_potsdam_7_9_RGB | dsm_potsdam_07_09 | top_potsdam_7_9_label |
| top_potsdam_7_10_RGBIR | top_potsdam_7_10_IRRG | top_potsdam_7_10_RGB | dsm_potsdam_07_10 | top_potsdam_7_10_label |
| top_potsdam_7_11_RGBIR | top_potsdam_7_11_IRRG | top_potsdam_7_11_RGB | dsm_potsdam_07_11 | top_potsdam_7_11_label |
| top_potsdam_7_12_RGBIR | top_potsdam_7_12_IRRG | top_potsdam_7_12_RGB | dsm_potsdam_07_12 | top_potsdam_7_12_label |
| top_potsdam_7_13_RGBIR | top_potsdam_7_13_IRRG | top_potsdam_7_13_RGB | dsm_potsdam_07_13 | |

Fig. 16: Details of all patched data images

Overview of the individual patches (with .tif extensions, all images have dimension 6000 x 6000 pixels). Filenames of TOP RGBIR: true orthophoto with four channels red, green, blue, infrared; TOP IRRG: true orthophoto with three channels infrared, red, green; TOP RGB: true orthophoto with three channels red, green, blue; DSM: digital surface model; GT: ground truth labels (empty if not made available to participants and used by the benchmark organizers for evaluation of the results). Note that no gaps exist between neighboring images tiles to enable models with context, better computation of normalized DSMs etc. The first number always indicates the row (top to bottom, i.e. north to south) and the second number the column (left to right, i.e. west to east). For example, direct neighbors of 5_12 are 4_12 (above), 6_12 (below), 5_11 (left), and 5_13 (right). All images come with corresponding .tfw-files that provide geo-reference in UTM WGS84 coordinates.

We choose the RGB, IRRG and DSM images for training. The image numbered 7_10 has error annotations, which will bring about the degradation of the network segmentation performance. So, it is removed from experiments. For the experiments in Section 5.4.1 and Section 5.4.2, the training set contains all remaining 23 images in Potsdam dataset. In Section 5.4.3 and Section 5.4.4, we divide the labeled Potsdam datasets into two parts, 18 images as the training set and 5 images (image numbers 7_7, 7_8, 7_9, 7_11, 7_12) as the validation set.